\documentstyle[12pt]{article}
\begin{document}

\vspace*{-1cm} %adhoc
\baselineskip=1.1\baselineskip

\begin{center}
{\Large \bf 
\mbox{Question Answering System Using Syntactic Information}\\}

{\large \bf 
Masaki Murata \hspace{0.6cm} Masao Utiyama \hspace{0.6cm} Hitoshi Isahara\\}

{\large
Intelligent Processing Section, 
           Kansai Advanced Research Center, \\
           Communications Research Laboratory\\
           Ministry of Posts and Telecommunications\\
           588-2, Iwaoka, Nishi-ku, Kobe, 651-2492, Japan\\
           TEL:+81-78-969-2181 \, FAX:+81-78-969-2189 \, http://www-karc.crl.go.jp/ips/murata\\
           \{murata,mutiyama,isahara\}@crl.go.jp}
\end{center}

\begin{abstract}
Question answering task is now being done in TREC8 using English documents. 
We examined question answering task in Japanese sentences. 
Our method selects 
the answer by matching 
the question sentence with knowledge-based data 
written in natural language. 
We use syntactic information 
to obtain highly accurate answers. 
\end{abstract}

%%%%%%%%%%%%%%%%%%%%%%%%%%%%%%%%%%%%%%
\section{Introduction}
%%%%%%%%%%%%%%%%%%%%%%%%%%%%%%%%%%%%%%
Question answering task has been done in TREC8 
using English documents \cite{trec8qa}. 
Here, we  
examine question answering task in Japanese sentences\footnote{With 
respect to Japanese sentences, 
domain-dependent work such as that on 
dialogue systems and help systems has been 
done \cite{kumamoto99B} \cite{higasa99B}, but  
little work has been done on detecting the answer from 
natural-language databases 
as in question answering task. 
However, much has been done on 
English sentences, 
such as work on 
detecting sentences in written answers \cite{morton99A} to 
work on detecting answers themselves \cite{MURAX}. }$^{,}$\footnote{This paper outlines one part of a question answering 
system that we have been developing for a long time \cite{murata94_QA_eng} \cite{murata99_QA_saki_eng}.}. 
Our approach is to use syntactic information. 

%%%%%%%%%%%%%%%%%%%%%%%%%%%%%%%%%%%%%%
\section{Question Answering System}
%%%%%%%%%%%%%%%%%%%%%%%%%%%%%%%%%%%%%%

\subsection{Outline}

\begin{enumerate}
\item 
The system detects keywords in question sentences, 
and then detects sentences in which 
the sum 
of the keywords' IDF values is high\footnote{In this paper, 
the system detects one sentence by one sentence. 
However, 
it would be better to detect a series of sentences and 
detect the answer from a series of sentences. 
If a series of sentences is used to detect the answer, 
context information can be used.}. 

\item 
The question sentences and the detected sentences 
are parsed by the Japanese syntactic analyzer \cite{KNP2.0b6_e}. 
(This allows us to obtain the dependency structures.) 

\item 
The answer is selected by matching 
a question sentence and the detected sentences 
using syntactic information. 
How this is done is described in the next section. 
\end{enumerate}

\subsection{Matching a Question and Detected Sentences Using Syntactic Information}

We use the syntactic information 
when matching a question sentence and the detected sentences. 

The score of a detected sentence $s$ is as follows. 

\begin{equation}
\label{eq:ruijido}
  Score(s) = B1(s) + \alpha * B2(s) - \beta * DNUM(s) 
\end{equation}

\begin{equation}
  B1(s) = \displaystyle \sum_{\begin{minipage}[h]{7cm}
      all bunsetsus b in the question sentence 
    \end{minipage}}  BNST1(b)
\end{equation}

\begin{eqnarray}
  B2(s) & = &\displaystyle \sum_{\begin{minipage}[h]{7cm}
      all pairs of 
      two bunsetsus 
      (b1, b2) in the question sentence, 
      where b1 depends on b2
      (i.e. b2 is the head of b1.)
    \end{minipage}}  BNST2(b1, b2)
\end{eqnarray}

Each of the bunsetsus in the question sentence can be paired with 
one of the bunsetsus in the sentence $s$ 
in order to maximize the value of $Score(s)$. 
(A {\it bunsetsu} in Japanese corresponds to a phrasal unit such as 
a noun phrase or a prepositional phrase in English.) 
$BNST1(b)$ is the similarity between 
the bunsetsu $b$ in the question sentence and 
the bunsetsu in the detected sentence $s$ 
paired with the bunsetsu $b$. 
$BNST2(b)$ is the similarity between 
the set of the two bunsetsus, ($b1$, $b2$), and 
the set of the two bunsetsus in the detected sentence $s$ 
paired with $b1$ and $b2$. 
$DNUM(s)$ is the number of the bunsetsus of the sentence $s$. 
$\alpha$ and $\beta$ are constants, 
and are set by experiment. 
(Although 
we use only monomial and binomial syntactic information in Eq. \ref{eq:ruijido}, we can also use trinomial or polynomial syntactic information.)

We calculate the similarity between two words 
by using the EDR dictionaries \cite{EDR93e}\footnote{
The similarity between words can be 
handled by using thesauri. 
But the similarity between long expressions 
such as clauses is difficult to handle. 
To solve this problem, we have already considered the method of 
using rewriting rules \cite{chomsky56}. 
This method will be described in later papers.}. 
In the case of the bunsetsu containing an interrogative pronoun,  
the similarity is calculated according to the situations. 
For example, 
when a bunsetsu in the question sentence is ``where'' 
and the paired bunsetsu in the sentence $s$ has
the meaning of location\footnote{Specifying 
bunsetsus whose meanings are locations 
is done by using thesauri such as the EDR dictionaries.}, 
the similarity between them is set to high. 

Our system performs the above matching process and 
selects the answer from the sentence having the highest score. 
The answer is selected by 
considering a bunsetsu paired with a bunsetsu 
containing an interrogative pronoun as the desired answer. 

In general, the answer of the question sentence can be 
obtained by matching the question sentence and 
the database sentences. 
In the case of YES-NO questions, 
the system has only to match the question sentence and 
the database sentence, and outputs YES if matched (or NO otherwise). 
In the case of fill-in-the-blanks-type questions\footnote{
The process of solving fill-in-the-blanks-type questions 
can be considered as a case of ellipsis resolution 
if the blanks are considered as ellipses. 
We have already discussed how 
corpora can be used in 
ellipsis resolution \cite{murata_anaphora_all_NLC_eng}. 
So we should be able to use corpora to fill blanks in 
the fill-in-the-blanks-type questions.}, 
the system has only to consider 
the element of the database sentence, paired with 
an interrogative pronoun such as ``What'' 
as the desired answer. 
Our approach here is an implementation of 
this idea using syntactic information. 

%%%%%%%%%%%%%%%%%%%%%%%%%%%%%%%%%%%%%%
\section{Example}
%%%%%%%%%%%%%%%%%%%%%%%%%%%%%%%%%%%%%%

This section shows three examples of 
when our system obtained correct answers. 

We used as question sentences 
the English-to-Japanese translations of sample sentences in TREC8. 
We used as database sentences 
the Daijirin Japanese word dictionary and 
the Mainichi Japanese newspaper (1991-1998). 
When we use the Daijirin dictionary, 
we added the strings 
``entry word $+$ {\it wa} ({\sf topic-marking functional word})'' 
to the beginning of each sentence. 

First, we inputted the following Japanese sentence into our system. 

\vspace{0.5cm}

\begin{tabular}[h]{lllllll}
{\it uganda} & {\it no} & {\it shuto} & {\it wa} & {\it doko} & {\it desu} & {\it ka}.\\
(Uganda) & (of) & (capital) & {\sf topic} & (where) & (be) & (?)\\
\multicolumn{7}{l}{(What is the capital of Uganda?)} \\
\end{tabular}

\vspace{0.5cm}

\noindent
As a result of calculating the score of Eq. \ref{eq:ruijido}, 
the following sentence in the Daijirin dictionary 
had the highest score  and 
``Kampala'' was correctly selected. 

\vspace{0.5cm}

\begin{tabular}[h]{lllllll}
{\it kanpara} & {\it wa} & {\it uganda} & {\it kyouwakoku} & {\it no} & {\it shuto} & {\it desu}.\\
(Kampala) & {\sf topic} & (Uganda) & (republic) & (of) & (capital) & (be)\\
\multicolumn{7}{l}{(Kampala is the capital of the Uganda republic.)} \\
\end{tabular}

\vspace{0.5cm}

\noindent
The score is calculated in the following. 
\begin{eqnarray}
Score & = & 9.7 \mbox{(Matching between ``Uganda'' and ``Uganda republic'')}\\
& + & 5.9 \mbox{(Matching between ``capital'' and ``capital'')}\nonumber \\
& + & 1.6 \mbox{(Matching between ``capital of Uganda'' and ``capital of Uganda republic'')}\nonumber \\
&   & ... \nonumber \\
& = & 17.2 \nonumber
\end{eqnarray}

Next, we inputted the following Japanese sentence into our system. 

\vspace{0.5cm}

\begin{tabular}[h]{lllllllll}
{\it magunakaruta} & {\it ga} & {\it tyouin} & {\it sareta} & {\it no-wa} & {\it nan} & {\it nen} & {\it desu} & {\it ka}.\\
(Magna Carta) & {\sf subject} & (sign) & {\sf passive} & {\sf topic} & (what) & (year) & (be) & (?)\\
\multicolumn{9}{l}{(What year was the Magna Carta signed?)} \\
\end{tabular}

\vspace{0.5cm}

As a result of calculating the score of Eq. \ref{eq:ruijido}, 
the following sentence in the Daijirin dictionary 
had the highest score  and 
``1215'' was correctly selected. 

\vspace{0.5cm}

\begin{tabular}[h]{llllllllllllllllllllllll}
{\it magunakaruta} & {\it wa} & {\it 1215} & {\it nen} & {\it igirisu} & {\it no} & {\it houken} & {\it shokou} & {\it ga} \\
(Magna Carta) & {\sf topic} & (1215) & (year) & (England) & (of) & (feudal) & (lords) &  {\sf subject} \\[0.2cm]
\end{tabular}

\begin{tabular}[h]{llllllllllllllllllllllll}
 {\it kokuou} & {\it jon} & {\it ni} & {\it semari}, & {\it ouken} & {\it no} & {\it seigen} & {\it to}  \\
 (king) & (John) & {\sf object} & (press) & (royal authority) & (of) & (limitation) & (and) \\[0.2cm]
\end{tabular}

\begin{tabular}[h]{llllllllllllllllllllllll}
 {\it shokou} & {\it no} & {\it kenri} & {\it wo} & {\it kakunin} & {\it saseta} & {\it bunsho}. \\
 (lords) & (of) & (right) & {\sf object} & (confirm) & {\sf causative} & (document). \\[0.4cm]
\multicolumn{24}{l}{(Magna Carta is the document 
in which feudal lords of England made King John}\\
\multicolumn{24}{l}{confirm the limitation of the royal authority 
and their rights in 1215.)}
\end{tabular}

\vspace{0.5cm}

\noindent
The score is calculated in the following. 
\begin{eqnarray}
Score & = & 32.0 \mbox{(Matching between {\it nan nen} ``what year'' and {\it 1215 nen} ``1215 year'')}\\
& + & 14.6 \mbox{(Matching between ``Magna Carta'' and ``Magna Carta'')}\nonumber \\
&   & ... \nonumber \\
& = & 48.1\nonumber 
\end{eqnarray}

Finally, we inputted the following Japanese sentence into our system. 

\vspace{0.5cm}

\begin{tabular}[h]{llllllllllllllllll}
{\it paakinson} & {\it byou} & {\it wa} & {\it nou} & {\it no} & {\it dono} & {\it bubun} & {\it ni-aru} & {\it saibou} & {\it no}\\
(Parkinson) & (disease) & {\sf topic} & (brain) & (of) & (what) & (area) & (in) & (cell) & (of) \\[0.1cm]
\end{tabular}

\begin{tabular}[h]{llllllllllllllllll}
{\it shi} & {\it ni} & {\it kankei-shite-imasu} & {\it ka}.\\ 
(demise) & (to) & (be linked) & (?)\\[0.3cm]
\multicolumn{18}{l}{(The symptoms of Parkinson's disease are linked to the demise of cells in what area} \\
\multicolumn{18}{l}{of the brain?)} \\
\end{tabular}

\vspace{0.5cm}

As a result of calculating the score of Eq. \ref{eq:ruijido}, 
the following sentence in the Mainichi newspaper 
had the highest score  and 
``substantia nigra'' was correctly selected. 

\vspace{0.5cm}

\begin{tabular}[h]{llllllllllllllllllllllllllllllllllll}
{\it paakinson} & {\it byou} & {\it wa} & {\it tyuunou} & {\it no} & {\it kokushitsu} & {\it ni-aru} & {\it meranin}\\
(Parkinson) & (disease) & {\sf topic} & (midbrain) & (of) & (substantia nigra) & (in) & (melanin) \\[0.2cm]
\end{tabular}

\begin{tabular}[h]{llllllllllllllllllllllllllllllllllll}
{\it saibou} & {\it ga} & {\it hensei-shi}, & {\it kokusitsu} & {\it saibou} & {\it nai-de} & {\it tsukurareru} \\
(cell) & {\sf subject} & (degenerate) & (substantia nigra) & (cell) & (in) & (be made) \\[0.2cm]
\end{tabular}

\begin{tabular}[h]{llllllllllllllllllllllllllllllllllll}
{\it shinkei-dentatsu-busshitsu} & {\it no} & {\it doupamin} & {\it ga} & {\it nakunari} & {\it hatsubyou-suru}, \\
(neurotransmitter) & (of) & (dopamine) & {\sf subject} & (run out) & (be taken ill) \\[0.2cm]
\end{tabular}

\begin{tabular}[h]{llllllllllllllllllllllllllllllllllll}
{\it to-sarete-iru}.\\
(be recognized)\\[0.4cm]
\end{tabular}

\begin{tabular}[h]{llllllllllllllllll}
\multicolumn{18}{l}{(Parkinson's disease is recognized when 
melanin cells in the substantia nigras of the midbrain}\\
\multicolumn{18}{l}{degenerate. 
The neurotransmitter dopamine, which is made in substantia nigra cells, runs}\\
\multicolumn{18}{l}{out, and Parkinson's disease arises.)}
\end{tabular}

\vspace{0.5cm}

The score was calculated in the following. 

\begin{eqnarray}
Score & = & 10.6 \mbox{(Matching between ``Parkinson's disease'' and ``Parkinson's disease'')}\\
& + & 6.3 \mbox{(Matching between ``cell'' and ``melanin cell'')}\nonumber \\
& + & 1.5 \mbox{(Matching between ``brain'' and ``midbrain'')}\nonumber \\
&   & ... \nonumber \\
& + & 0.4 \mbox{(Matching between ``area of brain'' and ``substantia nigra of midbrain'')}\nonumber \\
& + & 0.3 \mbox{(Matching between ``in area'' and ``in substantia nigra'')}\nonumber \\
&   & ... \nonumber \\
& = & 32.2\nonumber 
\end{eqnarray}

``cells in {\sf interrogative pronoun} of brain'' and 
``cells in substantia nigra of midbrain'' were matched, 
and ``substantia nigra'' was correctly selected\footnote{
It would be good to 
modify Eq. \ref{eq:ruijido} in order to increase 
the similarity of bunsetsus around an interrogative pronoun.}. 

%%%%%%%%%%%%%%%%%%%%%%%%%%%%%%%%%%%%%%
\section{Conclusion}
%%%%%%%%%%%%%%%%%%%%%%%%%%%%%%%%%%%%%%

We have outlined 
our question answering system using syntactic information. 
We intend to run more experiments, 
to make our system more robust. 

We think that 
the human sentence-reading process involves a 
matching process between the sentence being read now 
and data recalled in the brain \cite{mental_model}. 
Our question answering system matches 
question sentences and 
sentences in its database, 
and may therefore provide some clues 
to shed light on the human reading process. 
Future work will involve extending this current 
work 
to work on the human reading process. 

%%%%%%%%%%%%%%%%%%%%
%%% Bibliography %%%
%%%%%%%%%%%%%%%%%%%%
{\small
\baselineskip=1\baselineskip
\bibliographystyle{plain}

}
\end{document}